# NEURONAL AUDITORY MACHINE INTELLIGENCE (NEURO-AMI) IN PERSPECTIVE

E.N. Osegi[1, 2*],.
[*1]Department of Computer Science, National Open University of Nigeria (NOUN), Abuja, Nigeria
[2]SURE-GP Ltd, Nigeria

**Abstract**

The recent developments in soft computing cannot be complete without noting the contributions of artificial neural machine learning systems that draw inspiration from real cortical tissue or processes that occur in human brain. The universal approximability of such neural systems has led to its wide spread use, and novel developments in this evolving technology has shown that there is a bright future for such Artificial Intelligent (AI) techniques in the soft computing field. Indeed, the proliferation of large and very deep networks of artificial neural systems and the corresponding enhancement and development of neural machine learning algorithms have contributed immensely to the development of the modern field of Deep Learning as may be found in the well documented research works of *Lecun*, *Bengio* and *Hinton*. However, the key requirements of end user affordability in addition to reduced complexity and reduced data learning size requirement means there still remains a need for the synthesis of more cost-efficient and less data-hungry artificial neural systems. In this report, we present an overview of a new competing bio-inspired continual learning neural tool – Neuronal Auditory Machine Intelligence (NeuroAMI) as a predictor detailing its functional and structural details, important aspects on right applicability, some recent application use cases and future research directions for current and prospective machine learning experts and data scientists.

*Correspondence: ([end.osegi@sure-gp.com.ng](mailto:end.osegi@sure-gp.com.ng)`)*

## 1. Introduction

Artificial intelligence (AI) has evolved over time with sufficient progress made in the past one or two decades. An important and popular branch of AI, Artificial Neural Networks (ANN) or simply neural nets has been a very influential solution technique in many disciplines both in academic and industry setting. In actual fact, ANNs are artificial neural machine learning systems that exploit the unique properties or characteristic operations that occur in real biological neurons. These artificial agents of bio-advancement exploit some important facts, principles or theories about the components that make up the human brain for general problem solving. At the basic level, they are characterized by the existence of a population of artificial neurons which serve as information (data) processing units and cooperate or co-exist in such a way and manner that leads to some meaningful representation of input states or conditions e.g. what the weather may look like, the next game strategy to adopt, the next sequence of operation in a machine production shop etc.

ANNs have been deployed in a variety of application domains. Some areas of successful applications include the field of pattern recognition such as handwriting, facial and fingerprint recognition; forecasting or prediction problems such as weather, stock price and electricity forecasts; symbolic regression programming. The reason for these successes has been attributed to the ability of neural nets to approximate a universal function (Funahashi, 1989; Hornik et al., 1989; Bishop, 1995). Indeed, such networks used training and testing set of examples from a universal input set to perform this very important functional task. Some examples of conventional neural nets techniques that follow this aforementioned strategy include the feed-forward neural nets (FFnets) and all variants thereof, Recurrent Neural Nets such as the Long Short-Term Memory (LSTM) and its equivalent – the Gated Recurrent Units (GRU), Polynomial fitting networks such as the Group Method of Data Handling (GMDH), Online-Sequential Extreme Learning Machines (OS-ELM), the Convolutional Neural Nets (CovNets) of Yann Le Cunn and others.

While these neural net techniques have been promising, they are not without their own shortcomings. For instance, the FFnets including their recurrent types have been found to suffer from the vanishing gradient problem when trained at deeper levels (Hochreiter et al., 2001); indeed the use of back-propagation (gradient descent) training adds a complexity issue to this problem. More recent recurrent and sequential learning techniques such as LSTM, GRU, OS-ELM, and GMDH suffer from the needless complexity, excessive hyper-parameter tuning and still often requirement of very large training datasets in addition to the challenge of insufficient neuro-biological plausibility making them incompatible with present real time system requirements. Indeed, this issue has led to researchers devising alternating learning algorithms and/or computational strategies that optimize their performances (Barak, 2017) or that resort to the use of more compatible (progressive learning) architectures (Fayek, 2017; Fayek, 2019).

The recently developed HTM technology (Hawkins et al., 2010; Hawkins et al., 2021) though promising with more biological interpretations of how neural networks should be still suffer from a needless hyper-parameter tuning, algorithmic complexity and often requirement for a time stamp as is observed in its recent implementations (Struye & Latre, 2020). Indeed, the HTM and its recent variants still fall short of a detailed biological interpretation of the brain indicating that this matter is still a major research area for future explorative studies and for which there might not be an end in sight.

Thus, as can be seen from the aforementioned issues, there is an obvious need for very simple but powerful unconventional neural network implementations that draw inspiration from the intelligent operation occurring in mammalian brain; this is not out of place with mammalian brain structures as not all mammals exhibit the same level of algorithmic detail in the brain as there are in cats, mice's or humans. Indeed, by basing the performance of neural-like solution models to those that evolve more simplified structures and not just simply quantitative measures as fitness accuracy or fitness quality, we draw a common ground for attaining more efficient neural architectures. Philosophically, this important requirement may be likened to the Occam Razor Principle (ORP) which suggests the simplest hypothesis as the good one (Mitchell, 1997). With respect to the plethora of neural architectures in existence, the winning hypothesis then becomes the simplest one among the available or known neural architectures that is generally applicable across a large number of physical and non-physical computing domains. Thus, as a matter of research urgency, computational neuroscience researchers need to consider this important fact to facilitate the discovery of more simple and system-efficient neural models that are not just cost-friendly, but data and user friendly as well (refer also in Kell & Dermott, 2019).

This research contributes to the field of constrained bio-inspired continual learning artificial neural computing systems by proposing a far simpler processing structure than that found in conventional and other similar neural techniques. The proposal is developed on the basis of recent findings about the intelligent operations that occur in mammalian auditory cortex.

Some important areas of interests reported in this paper include the primitive intelligence and cognitive capability due to Mismatch Negativity (MMN) effect observable in mammals (Näätänen et al., 1978; Näätänen et al., 2001; Näätänen et al., 2007), functional re-organization ability in mammalian auditory cortex (A1) (Sollini et al., 2018) and sparse feature connections observable in cochlea neuron networks (Webster et al., 1992; Schuknecht, 1974).

## 2. A New Neural Predictive Technique Inspired by Auditory Processing

In order to harness effectively and realistically the intelligent operations that occur in mammalian brain, an in-depth understanding of the structure of a variety of mammalian brains is needed. In particular, it is expected that implementations of such feature a close connection and interactivity between the various replicas of the brain function. However, in practice this is not always possible resulting in a variety of constrained neural-like or brain-like artificial representations of the mammalian brain. One approach in this regard is to exploit the core operations that occur in the mammalian auditory system. The mammalian auditory system and in particular, the auditory cortex, is a unique and important part of a any mammal allowing the optimized processing of intelligible sound waves (signals) and featuring some very interesting functions as .sparse connectivity (Webster et al., 1992), functional re-organization (Sollini et al., 2018) and primitive intelligence (Näätänen et al., 2001).

In this section, we firstly introduce some inspired theories of the auditory cortex in the mammalian brain providing real world evidences based on sound and recent neuroscience discoveries. In subsequent sub-sections, we present our approach which is inspired by neural activity in the cochlea – a key part of the inner ear and the mammalian auditory system, and auditory sensations due to the mismatch negativity (MMN) effect occurring in human and animal subjects. Then we describe systematically, the process involved in designing a Neuro-AMI learning system and some current areas of applications.

## 2.1 Neural Activity in the Cochlea and Sparse Connectivity

The Cochlea (Cochlea Nuclei) represents an interesting and very important component part of the mammalian central auditory system. They represent the first relay center in the central auditory system It is located in the anterior part of the labyrinth (osseous labyrinth of the inner ear) just below the malleus (Clark, 2003, see Figure 1) and performs an important function of converting auditory sensations to sparse active neurons. Action potentials are typically generated in cochlear branch of the auditory central nervous system and this helps to carry information towards the outer cortex and other peripheral regions (Bess & Humes, 2008).

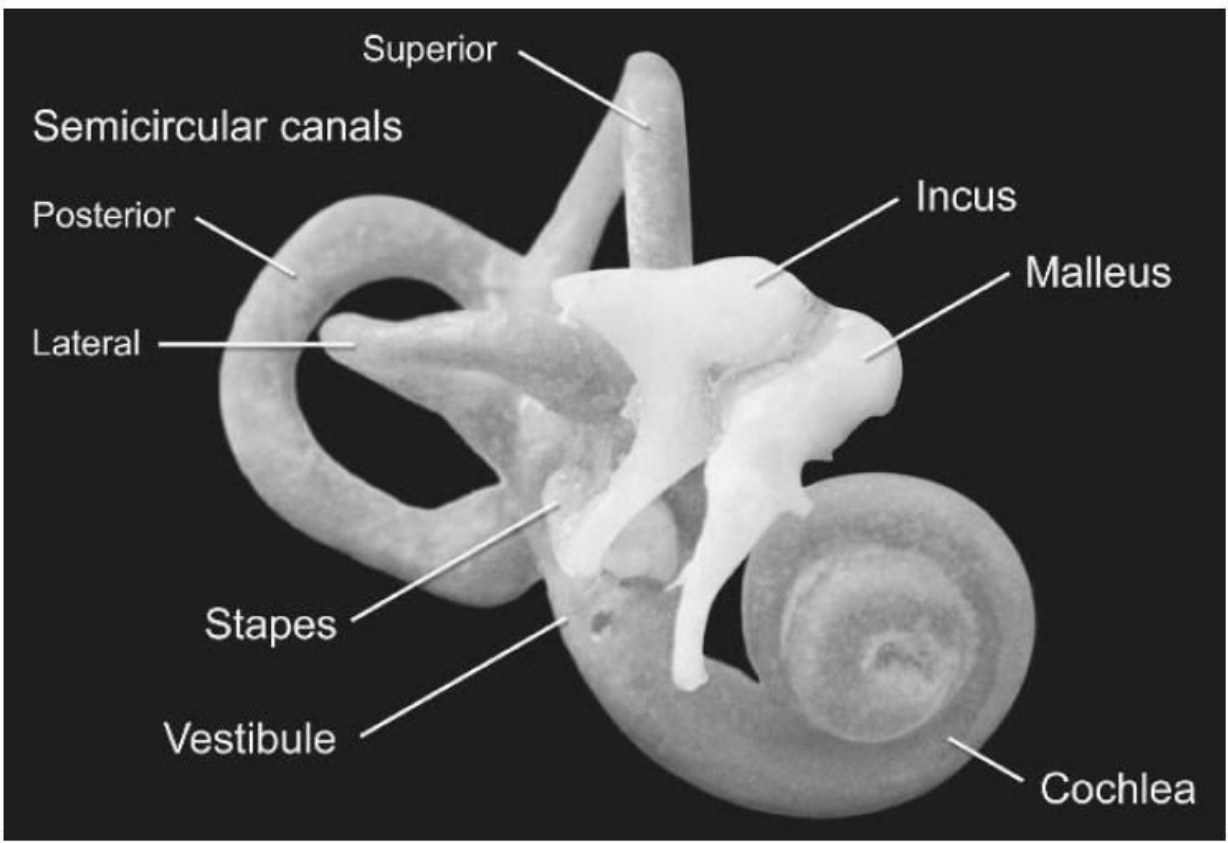

**Figure.1: Location of the Cochlea within the Middle to Inner Ear (Source: Clark, 2003).**

In several research studies, the Cochlea have been shown to exhibit sparsely connected hair-like frequency-tuned cells (Webster et al., 1992). Neurobiological explanations of encoding in mammalian organ (sensory) can also be found in the studies about the Cochlea as reported in (Webster et al., 1992; Schuknecht, 1974) in (Purdy, 2016). The structural operation may be seen from dorsolateral view including the Anterior Ventral Cochlear Nucleus (AVCN), Dorsal Cochlear Nucleus (DCN), Posterior Ventral Cochlear Nucleus (PVCN) and dual spiral ganglion nerve fibers (VIII N) as shown in Figure 2.

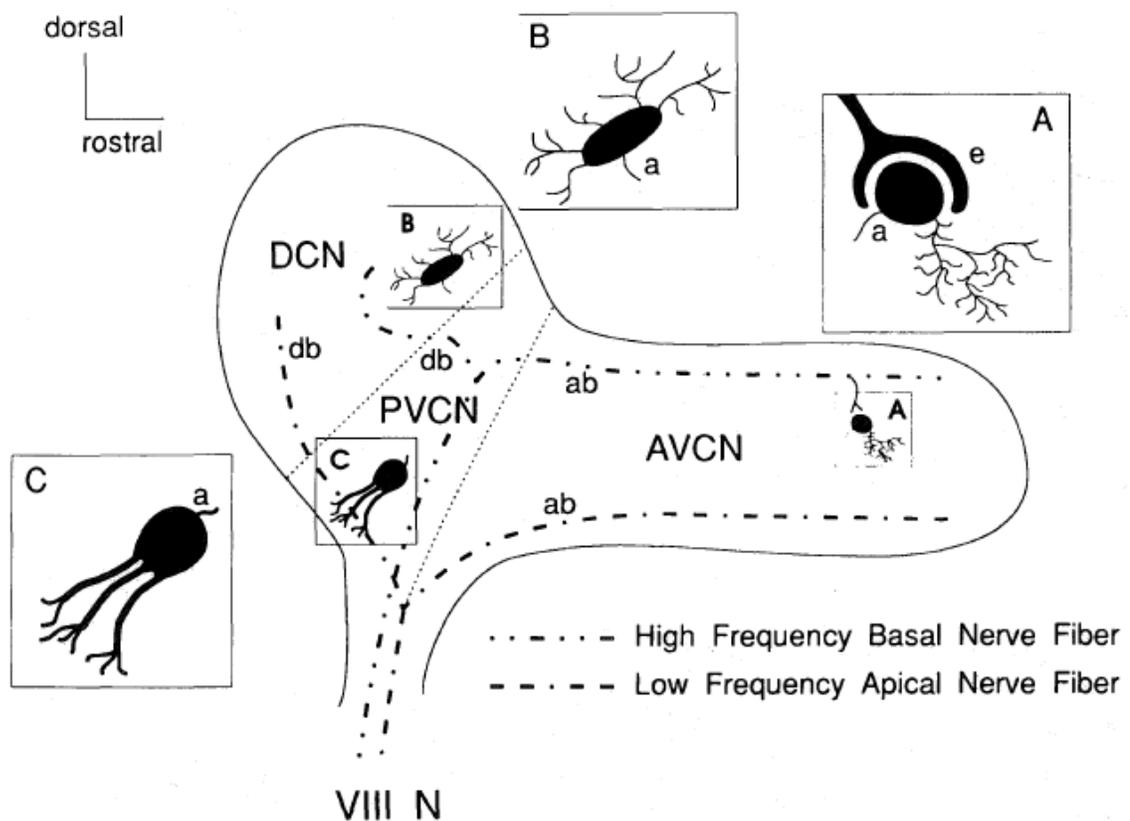

**Figure.2: Dorsolateral view of the Human Cochlear Nuclear Complex (Source: Webster et al., 1992).**

Typically, an axon with its associated end-bulb, synapse on a nucleic body of the neuron, *A* (see Figure 2) for a typical Anterior Ventral Cochlear Nuclei (AVCN). The interaction of more than one of these neurons (say several versions of *A*) leads to the formation of a neural network. In an Artificial Neural Network (ANN), we attempt to mimic the aforementioned operation which is albeit, an incomplete version of this feature or process using mathematical and/or logical functions.

AVCN neurons are specifically characterized by bushy cells (Brawer et al., 1974; Webster & Trune, 1982) and in conjunction with type-1 spral ganglion cells (neurons), they form very large end structures referred to as the end-bulbs of Held (see Ryugo & Fekete, 1982). These large amounts of synapses in the AVCN allow the highly efficient propagation of action potentials releasing enough neurotransmitters for post-synaptic depolarization (Webster, 1992).

Frequency tuning mechanisms have also been reported in some quarters (see Masterton et al., 1975) and its importance equally demonstrated in back propagation networks (Bogacz & Giraud-Carrier, 2000). Thus the Cochlea Nuclei (CN) can be seen as a sparsely tuned memory compressor. This idea may be exploited as a key ingredient to encode sensory input data streams for processing in a synthetic neuron model.

In the proposed NeuroAMI model implementation (see Appendix B for architectural details), we exploit the neuron sparse connectivity feature idea in two parts:

- In the first part we apply it to the modeling of the encoding logic for synthesizing compressed representations into a set of class-type neuron signals. Compressed representations have been reported to play a particularly important part in the learning of relevant features during atypical events (Schmidhuber, 1992).

- In the second part, these class-type neuron signals are used as inputs to a learning system that adaptively and temporally forms a sparse overlapping population of mean deviant weights. This second functionality greatly facilitates both single and multiple predictions akin to that obtained from the k-Winner Take All (kWTA) columnar representations or reference frames (Hawkins & Blakeslee, 2004; Hawkins et al., 2019).

With respect to item 2 in the aforementioned paragraph, it may be stated that the NeuroAMI model may be likened to a Candidate-Elimination (CE) strategy where the worst hypothesis (mean deviant weights) are excluded in the solution process (Mitchell, 1997). This may be achieved by temporally adjusting the hypothesis with respect to the computed prediction error states that meets a minimum satisfiability criterion.

We also exploit the idea of bushy AVCN neuron operations to validate the generation of a large population of mean deviant weights (typically 1000units) for performing artificial neuron activation operations.

At any point in time only a sparse set of these units will be activated in accordance with the prediction errors and the Hebbian Learning Rule (HLR).

**2.2. The Mismatch Negativity Effect (MMN) As Proof of Intelligent Processing**

Naatanen et al (2007) provided a detailed overview of the intelligent processing of auditory cortex in the context of the deviant stimulating effect provided by the MMN and particularly its ability to generalize has been validated and proven in related research aka "The Primitive Sensory Intelligence" (Näätänen et al., 2001).

Several important facts that may be identified are as stated below (Näätänen et al., 2007):

- Formulating important Concepts

- The extraction of rules
- Stimulus anticipation

Considering item 1, it is imperative to emphasize here that useful concepts can only be derived from useful situations. In this regard, MMN machine intelligence systems (MISs) have to be deployed in such a way as to only determine the essential parameter values that solve the problem; this follows from the importance theory of minimization of error bounds in an objective manner as opposed to prejudiced scenarios.

For item 2, the ability of MMN enable MISs to arrive at a consensus at which model bests describes a particular situation of event(s) makes such systems very useful at decision making expert applications – a number of such applications will be provided in later sections,

Stimulus anticipation has been an important requirement in many process and interpreted in a variety of ways. As a matter of fact, this requirement is often taken for granted in many businesses as is found in several organizations of developing economies but this is a really important feature of humans for efficient and reliable progressive business behaviors to be attained.

### 2.2.1. Neuronal Operations bordering on Change Detection

The observation of deviations from normality is not entirely a surprising operation performed by intelligent agents but how best to interpret such behaviors in AI based systems remains a core issue till the present moment. For the purposes of this study, the approach adopted in (Lieder et al., 2013) which states the following important response functions (Osegi & Anireh, 2020):

- MMN indexes only when a definite change occurs or not

- MMN indexes an unsigned or absolute value of the change in a physical parameter of any sensory input signal.
- MMN also indexes the signed value of the change in a physical property of a sensory input signal.

All these properties may be implemented by the software designer as is so desired and as the problem situation demands.

### 2.2.2. Neuronal Operations bordering on Model Adjustment

The model adjustment principle is one rule that fundamentally defines the way intelligent agents should be built. Why there is no generally agreeable rule to define how a model should be modified to solve a particular systems problem, we approach this issue based on the phenomenon of mismatch operations as elicited by a population of neurons fulfilling the MMN effect (Lieder et al., 2013).

## 2.3. The Neuronal Auditory Machine Intelligence Learning Mechanism

The approach based on Neuronal Auditory Machine Intelligence (Neuro-AMI) employs a Hebbian style of learning in a purely deterministic and temporal-adaptive manner considering the modification of a population of deviant mean weight units (Osegi & Anireh; 2020). When we compare this approach to conventional neural schemes, we see that a NeuroAMI learning model do not require an explicit random perturbation but an update based wholly on additive (reward) or subtractive (penalization) Hebbian rules have been used instead. Indeed, in neural selective attention systems, all but the most vital stimuli (deviant mean weights in our case) for a given observation task are filtered out (Bundesen, 1996; Dayan et al., 2000).

It is also very possible to implement multiplicative (amplifier rule) and divisive (divide and conquer rule) strategies to enhance the efficiency of the prediction. This has been practically validated in software by appropriate modification of learning rules in the most up to date NeuroAMI program code. The learning rule proposed in this context is as provided in Algorithm 1:

**Algorithm 1.** AMI Learning Algorithm

*i: Initialize $S_{pred}$, as prediction parameter, $S_{stars}$, as input expectation sequences (standards) state, $S_{dev(mean)}$ as deviant mean, $k_{adj}$ as population of rewards or penalties, $S_{diff(1)}$ as absolute difference between $S_{pred}$, and $S_{stars}$, $l_p$ as correction factor or bias.*
*ii:      for all $s \in s.S_{stars}$ do*
*iii:     if $S_{diff(1)} > 0$*
*iv:      $S_{dev(mean)} \leftarrow \left(S_{dev(mean)} - k_{adj}\right) \vee \prod 1/\left(S_{dev(mean)}, k_{adj}\right)$ %Weaken by subtractive or divisive penalty*
*v:      elseif $S_{diff(1)} < 0$*
*vi:     $_{dev(mean)} \leftarrow \left(S_{dev(mean)} - k_{adj}\right) \vee \prod \left(S_{dev(mean)}, k_{adj}\right)$ %Reinforce by additive or multiplicative penalty*
*vii:     else*
*viii:    $S_{dev(mean)} \leftarrow S_{dev(mean)} + l_p$*
*ix:     end if*

## 2.4. The Encoder

The encoder plays a valuable role in the development of machine intelligence systems as they ensure that data is captured in the right and most appropriate format for pattern analysis or further processing. As in humans, the interpretation of sight for instance should be captured using the relevant encoding unit, in this case, the eye. Similarly, the ear is responsible for encoding the sound waves and so on.

Neurobiological explanation of encoding in mammalian organ (sensory) can be found in the studies about the cochlea - the organ responsible for converting auditory sensations to sparse active neurons (Webster et al., & Schuknecht, 1974) in (Purdy, 2016). This idea is a key ingredient in our argument for the need for encoder-decoder component in today's neuron models.

Whatever the type of encoder used, the two key features desirable for implementation purposes are:

- Nature of Signal Type e.g. continuous streaming, static, etc
- Source determination and or representation e.g. symbolically, textual, graphically etc.
- The range of the encoding process and how the representation may be decoded.

While quite a number of encoders exist, we have decided to use a novel implementation in our study that exploits the symbolic or textual nature of data i.e. we assume that all data forms whether they be streaming or fixed, pictorial or audio, exist as a set or sequence of symbol groups. This conceptual approach affords us the ability to transcribe a data source into symbolic representation in a universal enough manner.

The processes involved in our novel encoding and compression scheme are set out as follows (Listings 1 to 3):

**Listing 1:** Encoding Process of the Neuro-AMI Encoder

1. Represent a data source as group of streaming time sequenced symbols – As an example, for sound or audio source this can be done using speech-to-text conversion systems and image sources as a mapping of pixels in terms of symbolic integers.
2. Text (symbols) reading using suitable programming language and conversion in conformance to appropriate ASCII set.
3. Transform symbols into numeric (integer-type) representation – see Listing 1 code snippets.
4. Scaling all inputs based on the maxima of the integer representations between 0 and 1; in our case we used the max-max scaling where all initial integer representations are divided by a two-process maxima operation of their initial representations.

5. Iterative comparisons using a unique swap-matching algorithm (code snippets as shown in Listing 2); this also includes the important steps of binarization and subsequent transformation to decimal integers.
6. Encoder Sensor-class memory formation using Listing 3. Note that an encoder class is formed first using an intuitive mathematical formula and the memories including all sequence group memories and the fundamental memory class structure are iteratively created.

**Listing 1:** Symbol-Integer Transformation Algorithm (SITA)

```
[roBo,coBo] = size(text_read_data);
for to_no = 1:roBo

kodevo=  double(cell2mat(B(to_no,:)));
lo_apno(to_no) = length(kodevo);
APstoredevo(to_no,1:lo_apno(to_no)) =  kodevo;

end
```

**Listing 2:** Swap-Match Algorithm (SMA)

```
uoo = APstoredevo./max(max(APstoredevo));
j = roBo;
fori = 1:roBo
A_j(i,:) = j;
A_i(i,:) = i;
        k1(i,:) = uoo(j,:);
        k2(i,:) = uoo(i,:);
Agn(i,:) = (uoo(j,:)==uoo(i,:));%% j == nth observation
%% also you can set j == 1st
%% observation during the encoding
%% run
%Agn_str(i,:) = num2str((uoo(j,:)==uoo(i,:)));
%Binarization step:
Agnn(i,:) = Agn(i,:) + 48;
Agn_str(i,:) = char(Agn(i,:) + 48);
%Binary-Decimal Integers Step:
Agn_str_dec(i,:) = bin2dec(Agn_str(i,:));

end
```

Using Listing 2, it is then possible to build a sensor-class temporary memory structure as shown in Listing 3:

**Listing 3:** Sensor-class Memory Algorithm (ScMA)

```
%% Encoder Class Formula:
Agn_scale = Agn_str_dec/max(Agn_str_dec);
Agn_class=  floor(class_level.^(Agn_scale));

%% Input the encoded class to input calling variable
input_data_nn = Agn_class;

%% Build default sensor class structure memory:
% Will be used for inference (decision-making) after prediction is
% deemed satisfactory ...
%
class_default = (1:class_level)';

fori= 1:length(input_data_nn)
%B_strings = (cell2mat(B(i,:)));
%len_B_end(i,1) = B_strings(length(cell2mat(B(i,:))));
for j = 1:class_level
%sensor_class(i,1) = find(input_data_nn(i,1)==class_val)
%sensor_class_id(i,1) = find(strcmp(B(i,1),class_val_symbolic))
if(Agn_class(i,1)== class_default(j,1))

sensor_class_memory(j,1) = (B(i,1));

end

end

end
```

## 3. Related Works
### 3.1. Hierarchical Temporal Memory

The Hierarchical Temporal Memory (HTM) for short, is an emerging neuro-computational technique that draw inspiration from the inter-operation of cortical columns and the concept of thousand brains (Mountcastle, 1997; Hawkins et al., 2010).

As evidenced in recent works such as in taxi passenger prediction (Cui et al., 2016), online anomaly detection in streaming data (Ahmad et al., 2017) and image recognition/classification tasks (Mattson, 2011; James et al., 2018), the HTM had shown superior or competitive performances with the state-of-the-art.

## 3.2. Spiking Neural Networks

The Spiking Neural Nets (SNNs) represent an emerging solution to the development of more bio-inspired software ANN based solutions to computational problems by incorporating more realistic versions of the conventional Feed-forward based ANN (FFANN) using especially designed spike timing pulsating units in both learning and activation functions (Awad & Khanna, 2015). These classes of ANNs have shown to follow the chemistry and physics associated with neural processing circuitry exploiting some so called membrane potential formation rules due to a spiking activity and sparsity (Maas, 2002; Eshraghian et al., 2022) or due to some sort of electronic (biophysical) representations as reported in (Tağluk, 2019, Tağluk & Isik, 2019).

In particular, it has been shown in (Tağluk & Isik, 2019), the frequency dependent characteristic of biophysical SNN leading to higher errors at high spike frequencies and lower errors at the lower spike frequencies. Indeed, we also seek to replicate this functionality in our proposed neuronal model by the use of a class-integer tuning parameter.

## 4. The Problem

The emerging artificial biological neural networks are faced with the utmost task of performing a reliable prediction task over a large number of domain independent applications. But this demands that such networks possess the universal computing capability that is clearly absent in many of the analogues of real brains. Thus, most research has focused on certain key areas of true neural processing amongst which include:

- The ability of neuron circuits to compute and store efficiently memory representations of input patterns in a continual learning manner
- The ability of neuron circuits to deal on a variety of input data
- The ability to deal with relatively small input data

- The need for implementing less memory or physical hardware intensive artificial neural systems.

The implication of the above mentioned problems is that there will still be a need for more versatile artificial neural architectures in the present time and near future. Thus, it is expected that research in the development of more resourceful business oriented ANNs will continue till we are able to effectively solve these pressing issues.

## 5. NeuroAMI Use Cases

In this section, some of the current applications of our proposed artificial neural model concept are provided and some examples equally provided using a dedicated NeuroAMI computer program. However, it is expected that more applications using variants of the NeuroAMI program will follow suit as more progresses are made in the design of auditory inspired artificial neural circuits.

### 5.1. One-Step Ahead Word and Character level Predictions

The first natural application of a NeuroAMI predictor is in making look-ahead estimates of group of symbols, texts or words. This is a very key area particularly in the field of natural language modeling such as in sentence prediction, intellisense based systems,

As an example, consider the task of predicting the following group of words:

*Car*
*Bus*
*Bus*
*Car*
*Car*
*Car*
*Car*
*Bus ...*

This sequence contains two key words, "car" and "bus"; but the bus came after the car at the end of the sequence so the look-ahead prediction should be the bus. However, as may be inferred from a critical examination, the car is more frequent than the bus so the NeuroAMI system predicts the car instead.

In Table 1 shows the encoded class field structure as found by the novel encoder scheme described earlier in section 2 (sub-section 2.4) while Table 2 shows the prediction class encoding pattern compared with expected class value. The computations were done considering a continual training set of 35% of the input encoded data, a deviant weight population of 1000units, a maximum deviant adjust factor of 2, and a class integer width of 5units (refer to Appendix A, Table A.2).

Table.1. Encoded class structure for Word Character Prediction Problem

| Representation Class | Word |
|---|---|
| 1 | Car |
| 2 | [] |
| 3 | [] |
| 4 | [] |
| 5 | Bus |

Table.2. Test Predicted vs. Expected class using sample data for 35% training data

| Predicted Class | Expected Class |
|---|---|
| 1 | 1 |
| 1 | 1 |
| 1 | 1 |
| 1 | 1 |

As can be seen from Table 1, the NeuroAMI assigns the Class 1 to "Car" and the Class 5 to "Bus". The square brackets denote redundant class cells which were not filled by the NeuroAMI encoder.

## 5.2. Forecasting physical parameters

One of the most popular uses of ANNs is in the forecasting of naturally occurring physical phenomena. Thus, NeuroAMI ANN model has recently been applied to distribution grid load forecasts (Osegi et al., 2020), temperature forecasts for class room air conditioning control (Osegi et al., 2021), transmission line fault signature prediction (Wokoma et al., 2022; Wokoma et al., 2023). The behavior during continual learning and hence predictions may be described by considering error response plots.

As an example, the continual prediction error response (mean absolute percentage error) using the NeuroAMI program for a gas pipeline pressure dataset is as shown in Figure 3.

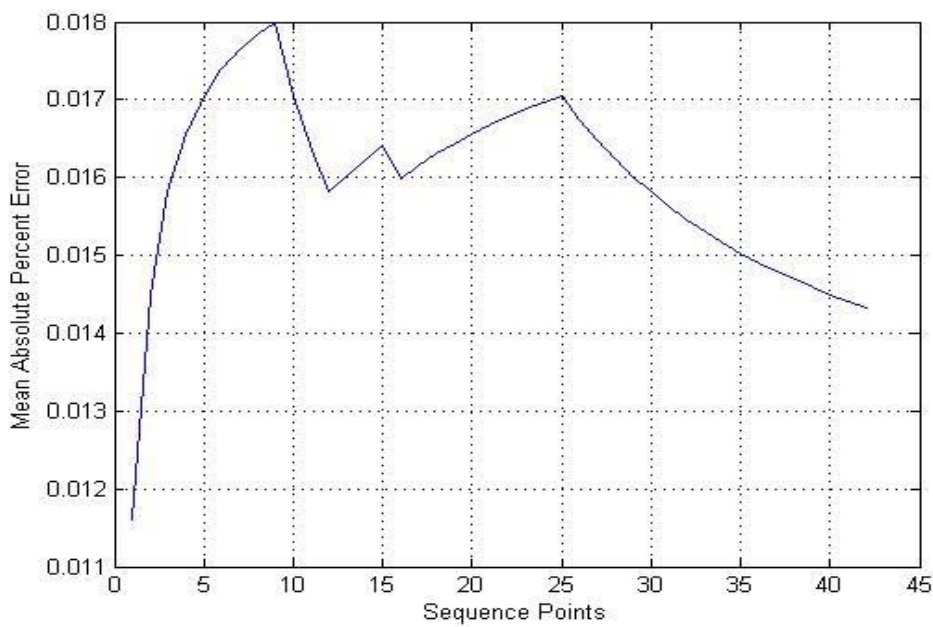

**Figure.3: Prediction Test Error Response for the Pipeline Pressure Dataset.**

The prediction describes the continual estimates on the test set (at 35% continual training set). As can be seen, the error margin increased at the start but gradually degrades over time towards the end of the data sequences. The class structure representations found by the NeuroAMI encoder is as shown in Table 3.

Table.3. Encoded class structure for Pipeline Pressure Prediction Problem

| Representation Class | Pressure (Bar) |
|---|---|
| 1 | 20 |
| 2 | [] |
| 3 | 15 |
| 4 | [] |
| 5 | 18 |

## 5.3. Consensus Modelling

More recently, the NeuroAMI ANN model has shown its unique ability in arriving at an agreeable operational expression model for gas turbine power output characterization based on a relatively limited dataset of symbolically evolved expressions (Osegi et al., 2023). This is achieved using the same encoding scheme as applied in the character level and pipeline pressure data simulations.

## 6. Conclusions and Future Research Directions

An overview of a new and promising technology and learning tool for predicting sensory events or observations is presented. This tool is primarily based on auditory machine intelligence inspired by recent discoveries about how the auditory cortex (A1) located in the brain might operate. A number of application scenarios are presented and discussed including the field of data encoding/classifications, time series forecasting and anomaly detection.

Though, the developed neural model proposed in this paper exploits a clearly deterministic process to synthesize a population of competing internally generated hypotheses, it is still possible to evolve these hypotheses using randomized weighting rules but it is recommended that researchers use the non-randomized much organic approach to ensure consistency in the generated results. Either way, it is possible that neural software designers may utilize a symbolic regression or genetic programming approach to compute the deviant mean weight for function-fitting tasks but this comes with the challenge of increased computational run-time.

The aspect of progressive learning and how to adapt to changing tasks or environments presents another more challenging requirement for the proposed neural technique. The proposed neural learning approach also presents an interesting but challenging area for possible applications in dynamical systems such as robot navigation and process control and in memory retention systems needed for real brain emulation tasks. This and other issues raised remain an important and open area for future studies.

**Acknowledgments**

The authors are grateful to the SurePay Foundations Group for providing source code support.

**Appendix**

**A. Interpretations of Neural Operations in Mammalian Brain**

1. Interpretations based on Neural Activation Potential (Activation Functions i.e. non-linearities)

- Standard ANN and all variants thereof e.g. Recurrent neural networks

2. Interpretations based on the presence of a set of overlapping proximal, distal and feedback dendrites biasing an OR activation function (based on union operation principle) - the HTM neural network.

3. Interpretations based on an interacting set of sparsely integer encoded deviant and standard stimulating and predictive model activations - Neuro-AMI neural network (proposed).

Thus far, as can be seen from the aforementioned neural schemes, the Neuro-AMI is distinguishable by its characteristic (implicit) sparse integer encoding and deterministic predictive operation. This makes it a unique neuro-computational model for replicating some core operations of mammalian cortex. In particular, some key properties of processing in auditory cortex such as described earlier (Webster et al., & Schuknecht, 1974 in Purdy, 2016) and in (Naatanen et al., 1978; Naatanen& ....2001; Naatanen et al., 2007) are replicated for performing a variety of machine learning and intelligence based tasks.

**A.1: Interpretations of the Conventional ANN:**

- Data encoding is simply a data normalization operation.
- Prediction (inference) is performed by summation of adjustable weights and biases and passing through a non-linearity (activation function).
- Learning is done by back-propagating errors.

**A.2: Interpretations of the HTM:**

- Data is encoded as SDR bits using a variety of especially designed encoders.
- Sensory Motor Prediction (inference) is performed using an overlap dot product matching rule between an input SDR at current time step and a set of internally synthesized SDRs derived from a union operation considering input SDRs at the previous time steps. Heavy use of reference frames (Hawkins et al., 2018).
- Learning follows Hebbian-type updates.
- Heavy use of Reference Frames (RFs) to represent different levels of abstraction; note that with RFs, it is possible to point to several unique identifiers (constants or variables) previously recorded in memory space so as to enhance the prediction experience.

**A.3: Interpretations of the Proposed NeuroAMI:**

- Data is sparsely encoded as whole number integers synthesized by a class-type signal non-linearity depicting shorter or longer wavelength periods (see Tağluk & Isik, 2019).
- Prediction (inference) is performed by temporarily computing a sparse population of weighted factors called the deviant means (see Algorithm 1) and adding the best fitted deviant mean to the data instance at the previous time step.
- Learning follows Hebbian-type updates.

**\*\*Notes:**

In the Neuro-AMI neural model, class-type symbolic and integer memory representations are formed where the integer memory representations are used for continual training and predictions and the symbolic representations for actual prediction decoding. It is important to emphasize here that depending on the class-signal value, the encoded representations can be more or less sparser. Indeed, sparsity is enforced by the tuning signal by converting the input data train into a sparse set of class neurons which is analogous to sparse behaviour in the cochlea (Webster et al., & Schuknecht, 1974) in (Purdy, 2016).

We hypothesize that intelligent predictions in the brain is the result of learning frequency tuned mismatch and reverse mismatch operations continually, by firstly encoding compressed versions of sensory input into unique integer classes automatically and adaptively predicting the class structures through time.

It may also be possible to exploit the feature of RFs as in the HTM neural networks in a NeuroAMI model. For instance, we might introduce an RF identifier as the last continual error state when the learning completes or stops i.e. prior to when inference (model testing) begins. In this regard, at the NeuroAMI inference model stage, it will have to adjust (update) its mean deviant weight if the computed new (test) errors falls below an automatically computed RF expectation value. Also, the RFs may be the basis of multiple final continual training errors (cte's) obtained from a corresponding set of winning overlapping columns. If a way is found to compress effectively, the input sequences coming into the brain's neural circuitry and encode it in highly constrained units of brain cells, then task of anticipating future events can be effectively reduced to a Class-Integer Prediction Problem (CIPP) akin to frequency depenedent error propagations observable in SNNs.

**Table A.1:** Neural Comparisons

| s/n | Neural Technique | Encoding | Learning Process | Learning Units Nomenclature |
|---|---|---|---|---|
| 1 | FFANN | Normalization to Real Numbers | Static | Random Weights |
| 2 | HTM | Binary | Continual | Random SDRs based on permanence |
| 3 | NeuroAMI (Proposed) | Integer-Class | Continual | Mean Deviant Weights |

**Table A.2:** The Core NeuroAMI Parameters

| s/n | Parameter | Default value (Typical Range) | Description |
|---|---|---|---|
| 1 | Max deviant adjust | 0.01 – 2.0 | Determines the precision range of the population of neurons |
| 2 | Deviant Neuron Population | 1000 | Generate a precise set of neural points for deviant-mean (dev) adjust |
| 3 | Class Integer Width | 2 – 10 | Frequency Tuning parameter for encoding |

**B. Architectural Structure of the NeuroAMI Artificial Neural Model**

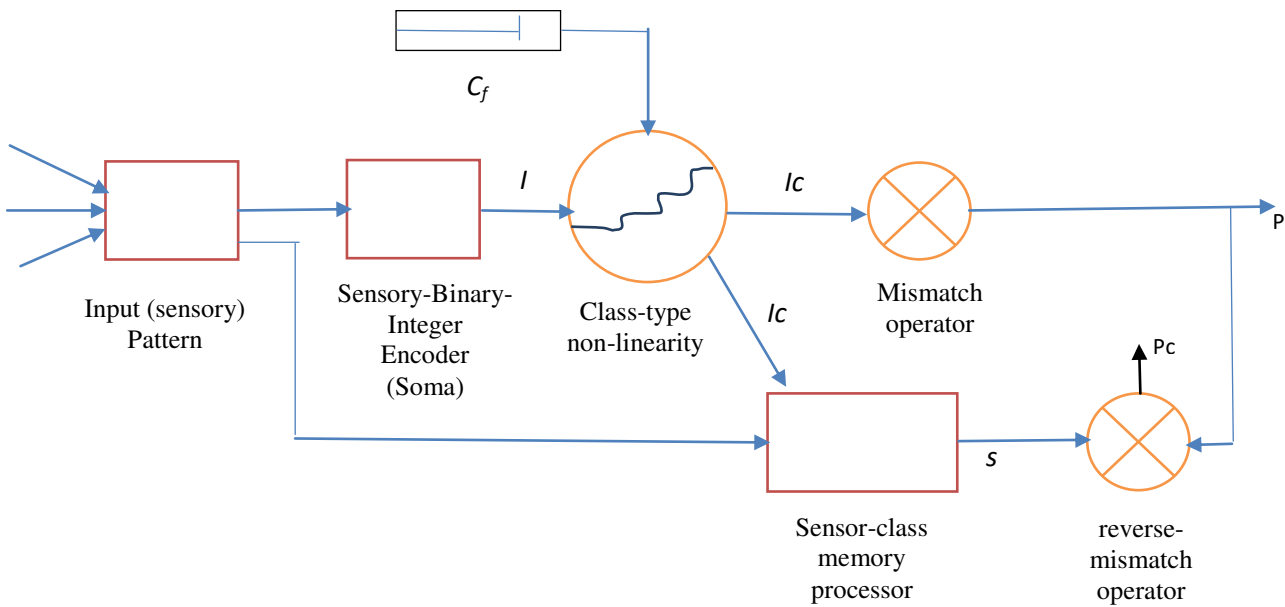

**Figure.B.1: Architecture incorporating both logical and functional components.**

$I_c$ – Integer class-type coding

P – Output Prediction

$P_c$ – Predicted cell (neuron)

s – Sensory memory cell (neuron)

$C_f$ – represents a frequency class signal level for tuning the class-type non-linearity

***Notes:**

The class type non-linearity allows for the generation of candidate cells (also called deviant neurons) via a mismatch and reverse-mismatch operation. The outputs are generated in mismatch operation step while the predicted output states are obtained in the reverse-mismatch operation step.